\documentclass[letterpaper, 10 pt, conference]{ieeeconf}  

\usepackage{newtxtext}
\usepackage{float}
\usepackage{placeins} 
\usepackage[hidelinks]{hyperref}

\IEEEoverridecommandlockouts                              

\usepackage{amssymb}
\usepackage{amsmath}

\usepackage{graphicx}
\makeatletter
\def\endfigure{\end@float}
\def\endtable{\end@float}
\makeatother
\usepackage[caption=false,font=footnotesize]{subfig}

\usepackage{cite}\usepackage{url}

\usepackage{multirow}
\usepackage{tabularx}
\usepackage{booktabs}
\usepackage{microtype}
\title{\LARGE \bf
HarvestFlex: Strawberry Harvesting via Vision-Language-Action Policy Adaptation in the Wild
}

\author{Ziyang Zhao$^1{^{,2}}$, Shuheng Wang$^{1}$, Zhonghua Miao$^{2}$ and Ya Xiong$^{1*}$
\thanks{This work was supported by the Haidian District Bureau of Agriculture and Rural Affairs, the BAAFS Innovation Ability Project (KJCX20240321), and the BAAFS Foundation for Excellent Young Scientists (Grant No. YKPY2025007). (*\textit{Corresponding Author: Ya Xiong, \tt\small yaxiong@nercita.org.cn})}
\thanks{$^{1}$The Intelligent Equipment Research Center, Beijing Academy of Agriculture and Forestry Sciences, Beijing 100097, China.}
\thanks{$^2$The School of Mechanical Electrical Engineering and Automation, Shanghai University, Shanghai 20044, China.}%
}

\begin{document}

\sloppy
\emergencystretch = 1em

\maketitle
\thispagestyle{empty}
\pagestyle{empty}

\begin{abstract}
This work presents the first study on transferring vision-language-action (VLA) policies to real greenhouse tabletop strawberry harvesting, a long-horizon, unstructured task challenged by occlusion and specular reflections. We built an end-to-end closed-loop system on the HarvestFlex platform using three-view RGB sensing (two fixed scene views plus a wrist-mounted view) and intentionally avoided depth clouds and explicit geometric calibration. We collected 3.71 h of VR teleoperated demonstrations (227 episodes) and fine-tuned $\pi_0$, $\pi_{0.5}$, and WALL-OSS with full fine-tuning and LoRA. Under a unified 50 trials real-greenhouse protocol and metrics spanning completion, $\pi_{0.5}$ with full fine-tuning achieved success rate of 74.0\% with 32.6 s/pick and damage rate of 4.1\%. Asynchronous inference-control decoupling further improved performance over synchronous deployment. Results showed non-trivial closed-loop picking with fewer than four hours of real data, while remaining limited by close-range observability loss and contact-dynamics mismatch. A demonstration video is available at: https://youtu.be/bN8ZowZKPMI.
\end{abstract}

\section{INTRODUCTION}

Strawberry harvesting is a high-value agricultural operation that still relies heavily on manual labor, leading to high seasonal labor demand, rising costs, and workforce instability \cite{10.1016/j.compag.2024.109468}. Unlike structured industrial settings, tabletop greenhouse strawberry farms pose substantial visual challenges (e.g., severe occlusions and a wide variation of illumination), while the fruit itself is delicate and highly sensitive to contact \cite{Xiong2020}. In practice, a full picking episode is a long-horizon, closed-loop sequence that typically alternates among target discovery, collision-aware approach, compliant engagement, detachment, placement, and reset; failures in any phase can damage the fruit or abort the task.

Under occlusions, specular reflections or exposure fluctuations, and state changes induced by contact, errors can accumulate and amplify along the perception--planning--execution chain \cite{rajendran2023towards}, resulting in limited robustness and poor transferability across farms. 

Traditional robotic fruit harvesting systems typically follow a modular perception--planning--control pipeline. A canonical setup consists of target detection and segmentation, pose estimation, motion planning, servo control, and a task-specific end-effector \cite{xiao2024review}. On the perception side, early approaches relied on hand-crafted color or shape features, whereas recent works increasingly adopt deep learning-based detectors and segmenters to identify fruits and estimate candidate picking points \cite{sun2026srrnet}. These visual outputs were then fused with depth detection and calibration to recover 3D target poses, followed by trajectory planning or visual servoing to execute approach and picking motions \cite{zhu2025_yolov11_skp}.

To cope with the delicacy of fruits and the occlusions of leaves and branches, many studies designed specialized end-effectors (e.g., soft grippers \cite{11231358}, suction devices, and cutting mechanisms \cite{sobol2024_headgrabber}) and incorporated compliant control strategies such as force or impedance control to reduce damage during contact \cite{ji2022_contact_impedance}. 

\begin{figure}[t]
\centering
\includegraphics[width=\columnwidth]{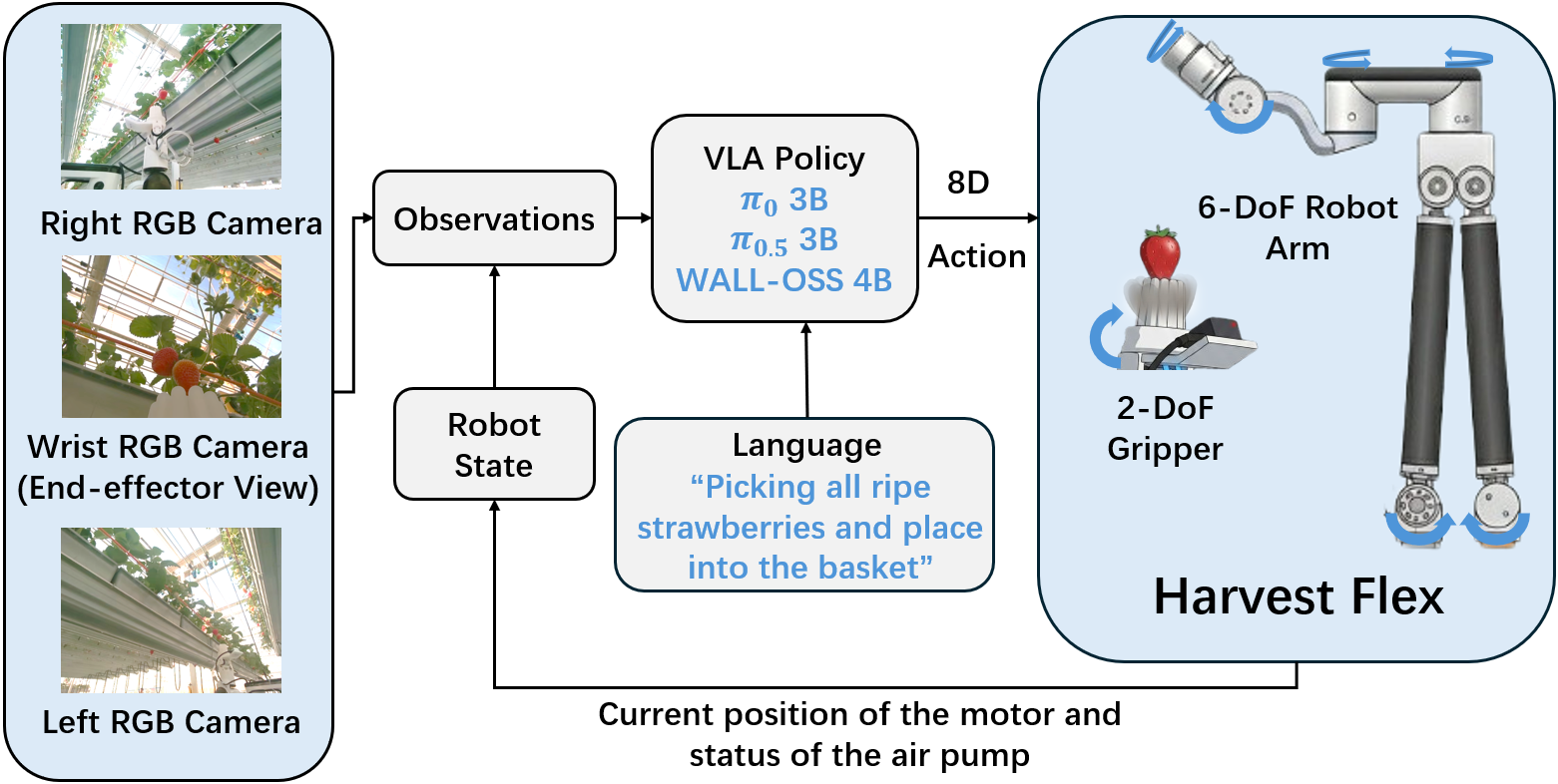}
\caption{HarvestFlex end-to-end closed-loop harvesting pipeline: three-view RGB observations and robot state, conditioned on a language goal, were mapped by a VLA policy to 8-D actions for real-world execution.}
\label{fig:pipeline}
\end{figure}

Although effective in controlled scenarios, these modular stacks often require substantial tuning for each farm and can still struggle when observations are partially missing or rapidly changing, motivating policies that can fuse rich visual context with task intent and react online \cite{ahn2022saycan}.

Recent progress in large-scale multi-modal pretraining has substantially advanced instruction-following robotic manipulation \cite{sapkota2026visionlanguageactionvlamodelsconcepts}. Meanwhile, some recent studies leveraged the Vision-Language Model(VLM) for subtasks such as ripeness recognition \cite{yang2025_agriGPTvl} and picking-path planning \cite{zhao2024_vlmpc}; for example, Ge et al. \cite{ge2025_occlusion_aware_sequence} proposed an occlusion-aware harvesting-sequence planner that used large language and vision-language models to generate cluster-aware pick orders by explicitly reasoning about visibility and reachability in densely occluded strawberry clusters. However, these systems typically do not directly output low-level control commands for robotic execution.

To bridge this gap, the community has increasingly explored how to introduce the notion of ``action'' into VLMs, giving rise to the VLA paradigm \cite{sapkota2025vision}. In VLA, a policy takes visual observations, natural-language goals, and robot state as input and directly predicts actions — a direction instantiated by RT-2 \cite{brohan2023_rt2} and OpenVLA \cite{kim2024_openvla} — thereby reducing the reliance on task-specific perception modules, explicit geometric modeling, and hand-crafted state machines. 

Representative open VLA baselines include $\pi_0$ \cite{pi0_2024}, which provided a reproducible end-to-end VLA training and deployment resource and demonstrates the potential of multi-task pretraining for transfer across embodiments and tasks. Building on this foundation, $\pi_{0.5}$ \cite{pi05_2025} further improved real-world stability and efficiency through engineering and training refinements. Finally, broader efforts such as WALL-OSS \cite{walloss_2025} targeted open-world interaction coverage and cross-scene generalization, validating their claims in more diverse environments and richer manipulation behaviors through architectural and curriculum-level advances.

However, systematic validation of VLA models on real-world fruit harvesting---a non-structured, contact-sensitive, long-horizon task---remains limited. So, we focus on three questions: (1) Are open-source VLA policies feasible for fruit-harvesting task, and what level of overall performance can they achieve? (2) What are the strengths and limitations of different VLA models and training strategies in terms of success, efficiency, and fruit damage? (3) Under the same training budget, does asynchronous inference improve detachment performance compared to synchronous inference?

In this work, the adaptation of the real strawberry-harvesting robot HarvestFlex (a 6-DoF arm \cite{chen2024design} with a 2-DoF compliant end-effector \cite{11231358}, project page: \url{https://xiong-lab.cn/}) was carried out and long-horizon real-world demonstrations were collected to fine-tune VLA models. The overall workflow proceeds as follows: collected demonstrations via VR teleoperation; adapted VLA models on the collected data; deployed the resulting policies for closed-loop real-world evaluation. The detailed closed-loop inference process is illustrated in Fig.~\ref{fig:pipeline}.

Our contributions are summarized as follows.

1) A closed-loop, end-to-end system for greenhouse tabletop strawberry harvesting has been presented and implemented, and the LeRobot \cite{cadene2024lerobot} data-collection framework has been integrated with the HarvestFlex robot, enabling continuous perception--decision--control execution in the real world.

2) Long-horizon demonstrations have been collected via VR teleoperation under diverse lighting and occlusion conditions, covering key stages of the picking workflow and providing a reproducible end-to-end data-collection recipe for contact-sensitive fruit harvesting.

3) A comprehensive evaluation protocol for VLA-based strawberry picking has been established, reporting success rate, cycle time, damage rate, typical failure scenarios, and stage-wise success, and including ablations on camera views and deployment settings under a unified protocol.

4) Under the above protocol, we systematically compared multiple open-source State-of-the-Art(SOTA) VLA models when transferring to a previously unseen embodiment, and studied the trade-offs between full fine-tuning and parameter-efficient fine-tuning, offering practical insights for real-world, non-structured fruit harvesting.

\section{METHOD}
Strawberry harvesting is formulated as a long-horizon, closed-loop manipulation task carried out in cluttered greenhouse tabletop environments.

\subsection{Task Definition and Stages}
\label{sec:task_definition}
In each episode, we split the behavior sequence into five stages:
(1) Target Selection: the robot identified a harvestable target (ripe and reachable) from the current view and produced a high-level intent for the subsequent approach.
(2) Approach \& Obstacle Avoidance: the end-effector was moved to an operable pose near the target ($<$1 cm) while avoiding contact with leaves and the rack; when necessary, the pose was adjusted to obtain a stable contact angle.
(3) Envelop \& Detach: the end-effector swallowed the strawberry and forms a stable suction, followed by snapping-ogg motions to detach the fruit from the stalk.
(4) Placement: the picked fruit was placed above the designated container and released to complete a successful pick.
(5) Retry or Reset: when a pick attempt failed, the end-effector immediately retried without homing. The system proceeded to placement only after a detached strawberry was secured in the gripper. The end-effector automatically reset to a safe pose for the next attempt only after a successful pick and placement.

Note that an episode may contain multiple such pick-and-place cycles to follow the given language instruction.

Stage labels were determined jointly by teleoperation logs and key event triggers, and were aligned to the image--action sequence via timestamps.

\subsection{System Setup}
\label{sec:system_setup}
We used HarvestFlex as our experimental platform. The end-effector was actuated by an air pump and employed a silicone structure to passively conform to the fruit during contact, thereby reducing bruising risk. The pump command was discretized and logged into three states (in, out, and idle) to simplify the action interface while retaining the key actuation modes required for harvesting.

\begin{figure}[t]
  \centering
  \includegraphics[width=0.9\linewidth]{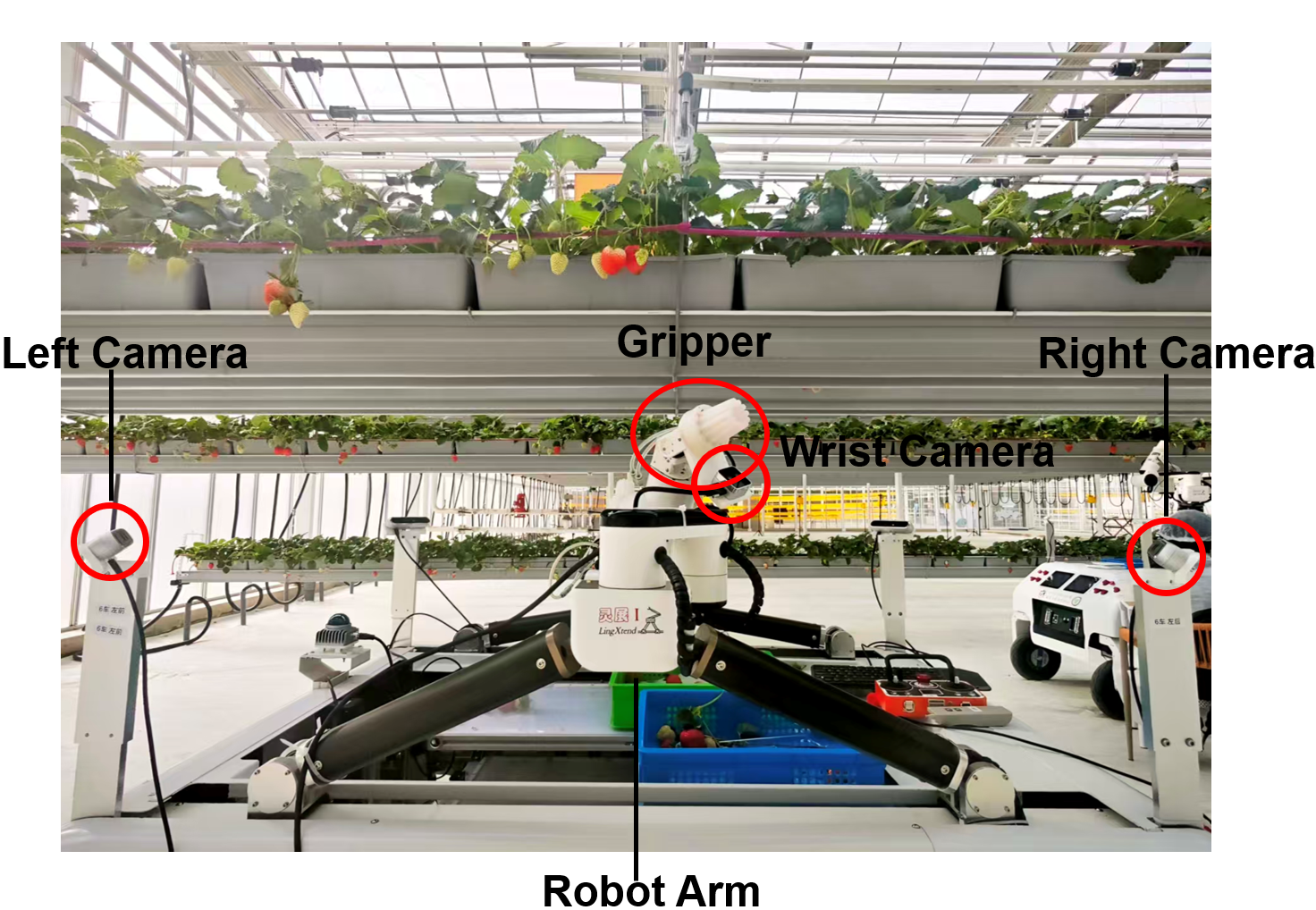}
  \caption{Real-world strawberry picking scene and camera configuration.}
  \label{fig:harvest_camera_picking_scene}
\end{figure}

The system was equipped with three cameras: two fixed scene cameras providing left and right views, and one wrist-mounted camera co-located with the end-effector. As shown in Fig.~\ref{fig:harvest_camera_picking_scene}, the left and right scene cameras were RealSense D455 (Intel, RealSense, USA), and the wrist camera was a RealSense D405 (Intel, RealSense, USA). To match the standard VLA input formulation and avoid additional engineering dependencies, only RGB images were used as visual input (i.e., no depth or point clouds and no explicit geometric calibration). This multi-view setup improved observability for both target selection and contact-rich manipulation, and mitigated view-specific degradation caused by occlusions and specular reflections.

The arm executed policy outputs through a velocity-mode interface, which alleviated motion jitter that could arise when policy actions were not perfectly smooth under position control. Low-level safety constraints were further enforced, including action clipping, workspace limits, and emergency-stop triggers, to prevent collisions and unintended damage to plants.

Hardware and data streams were integrated into the LeRobot framework, unifying multi-camera acquisition, timestamp alignment, policy inference, and command execution. The system also recorded reproducible logs of observations, actions, and key events for offline analysis and replay.

\subsection{Data Collection}
\label{sec:data_collection}
To cover both the global context needed for target search and selection and the local cues required for contact and detachment, a three-view layout was adopted. The scene cameras offered a wide field of view and stable context for selecting feasible targets, planning approaches, and resolving occlusions through viewpoint redundancy. The wrist camera provided high-resolution local observations at close range, reducing uncertainty due to end-effector occlusion and small pose errors during contact.

\begin{figure}[t]
    \centering
  \subfloat[]{\includegraphics[width=0.32\linewidth]{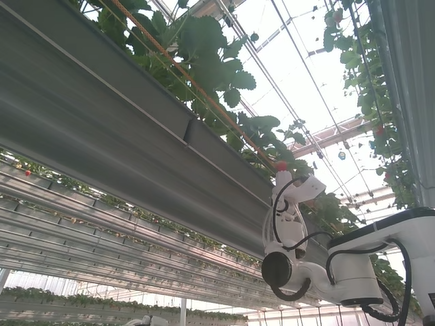}}
  \hfill
  \subfloat[]{\includegraphics[width=0.32\linewidth]{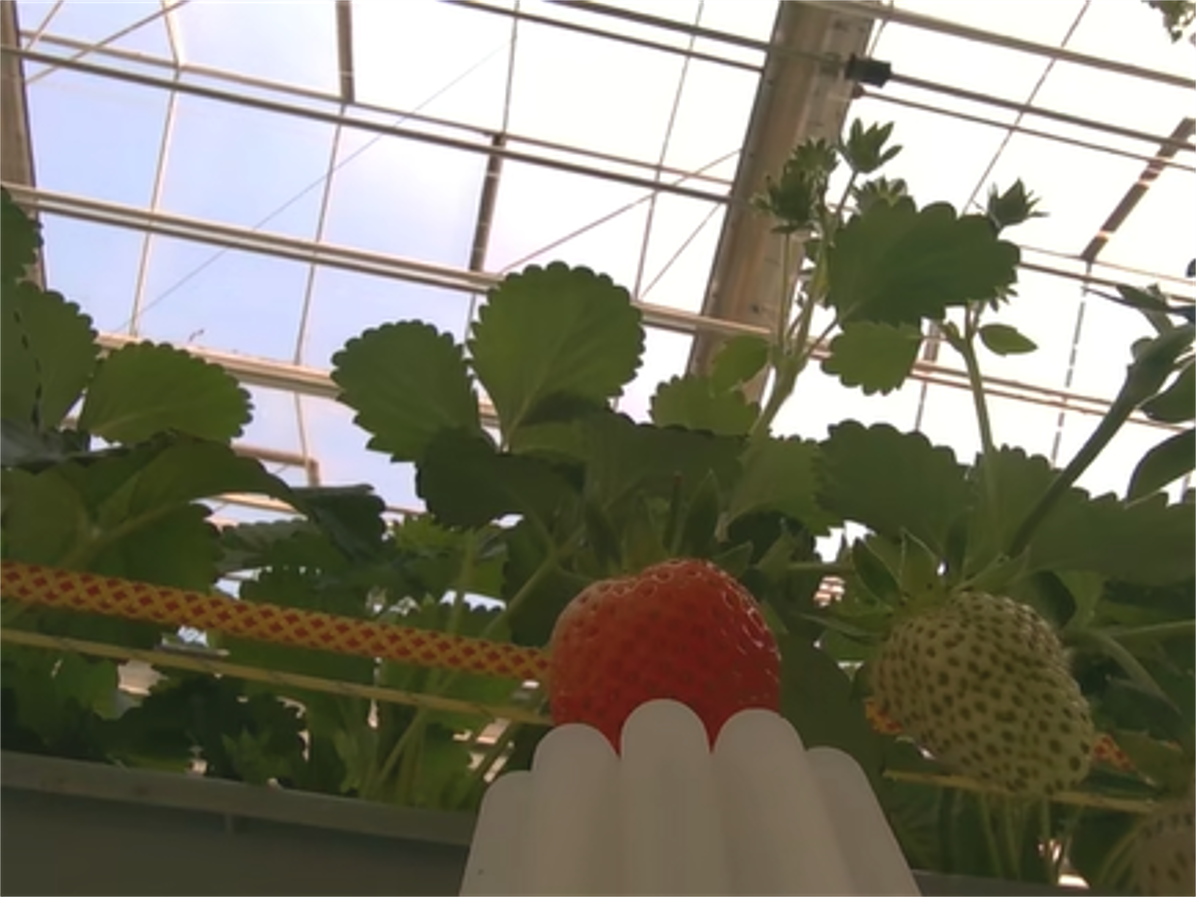}}
  \hfill
  \subfloat[]{\includegraphics[width=0.32\linewidth]{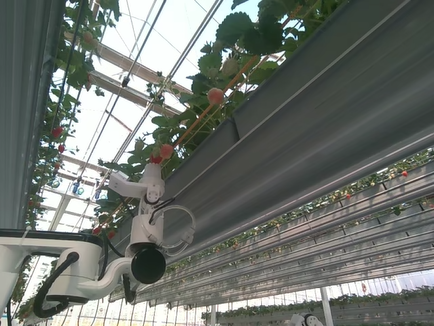}}
  \caption{Example three-view observations in our dataset: (a) left scene camera (D455), (b) wrist camera (D405), and (c) right scene camera (D455).}
  \label{fig:data_three_view_examples}
\end{figure}

Acquisition settings were standardized for each camera and key metadata were recorded for reproducibility. Details are summarized in Table~\ref{tab:camera_settings}, including RGB resolution and frame rate, auto-exposure and auto white-balance policies.

\begin{table}[t]
\centering
\caption{Camera acquisition settings and coverage.}
\label{tab:camera_settings}
\begin{tabularx}{\columnwidth}{l l X l}
\toprule
Camera & Model & RGB (W$\times$H) @ fps & Exposure / WB \\
\hline
Left scene  & D455 & 1280$\times$720 @ 30 & Auto / Auto \\
Right scene & D455 & 1280$\times$720 @ 30 & Auto / Auto \\
Wrist       & D405 & 640$\times$480 @ 30  & Auto / Auto \\
\bottomrule
\end{tabularx}
\end{table}

Demonstrations were collected using a lightweight VR teleoperation setup (Meta Quest3, Meta, USA). The operator received a first-person visual feedback stream in the headset and used handheld controllers to generate a continuous stream of end-effector pose commands, enabling smooth teleoperation of HarvestFlex.

\begin{figure}[t]
  \centering
  \subfloat[third-person view]{\includegraphics[width=0.48\columnwidth]{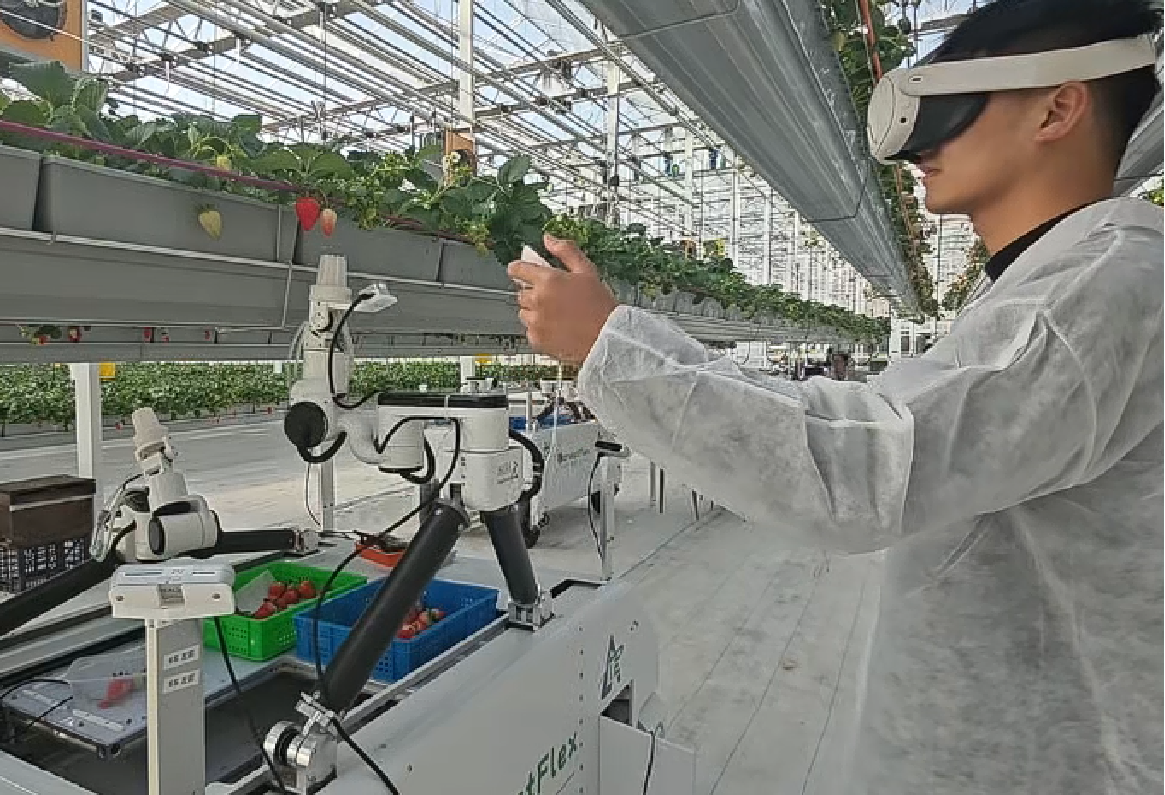}} \hfill
  \subfloat[first-person view]{\includegraphics[width=0.48\columnwidth]{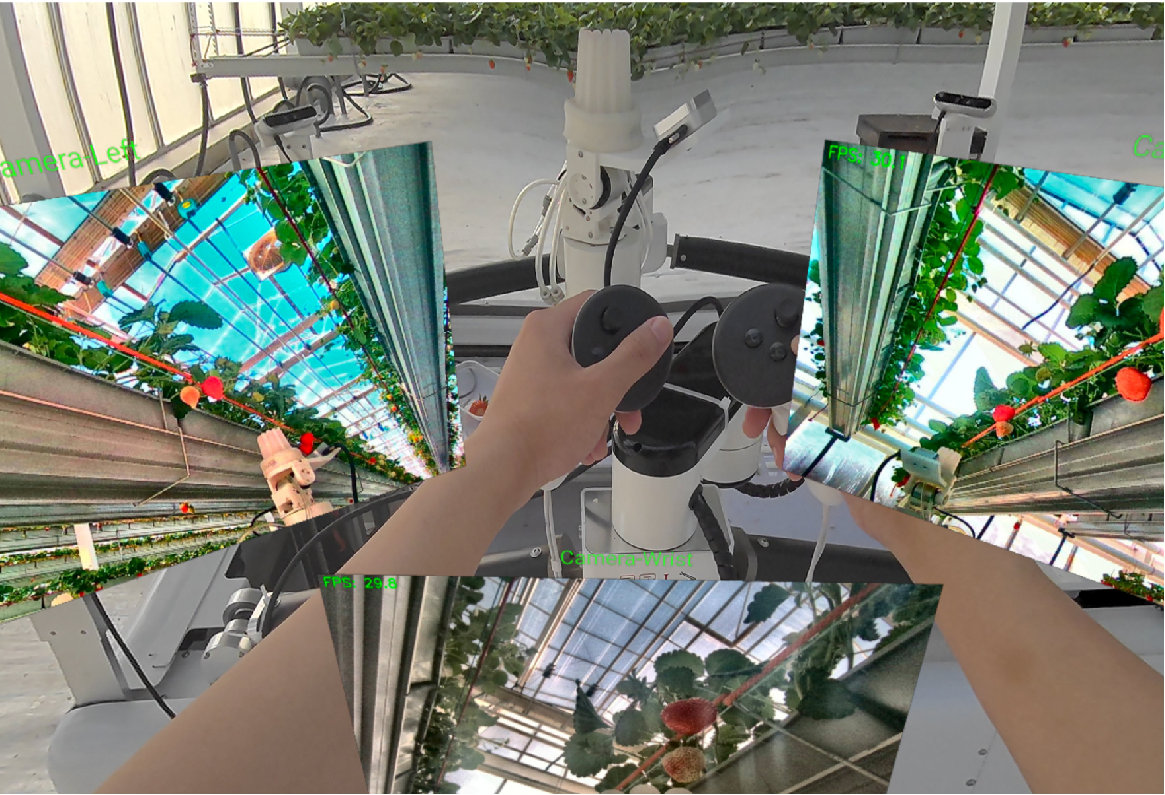}}
  \caption{VR teleoperation interface and single-operator control scheme. (a) Schematic of the third-person viewpoint used to illustrate the teleoperation setup. (b) First-person view from the operator's headset recorded during demonstration collection.}
  \label{fig:teleop_interface}
\end{figure}

Multi-camera images were streamed to the VR device with low latency for real-time operation, while being recorded in parallel for offline training. The system synchronously logged observations, actions, control commands, key events (pause/resume/stop), and timestamps to support deterministic replay and alignment. Fig.~\ref{fig:teleop_interface} shows the teleoperation scene and the corresponding first-person view during data collection.

\begin{figure}[t]
  \centering
  \includegraphics[width=0.6\linewidth]{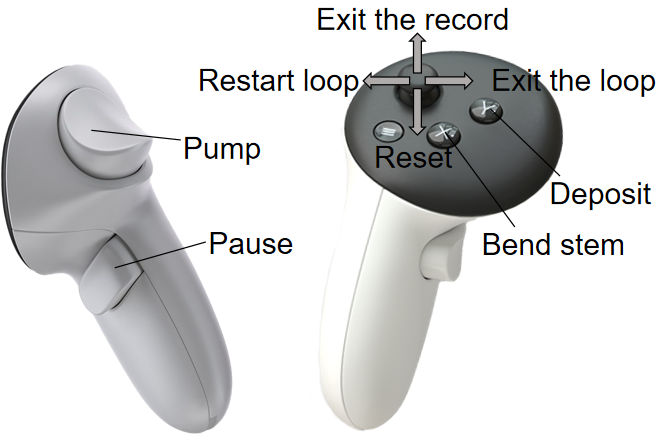}  
  \caption{VR controller button map for one-person operation.}
  \label{fig:vr_single_operator_control}
\end{figure}

To reduce demonstration cost and support long-horizon sessions, session control was implemented via controller interactions, allowing a single operator to manage the full data-collection workflow. Given the fragility of both the robot and the crop, one-click pause and safe-stop functions were provided. When resuming from a paused state, target commands were interpolated to avoid discontinuous jumps that could induce large instantaneous velocities. The detailed controller button map is shown in Fig.~\ref{fig:vr_single_operator_control}.

All modules ran above 50~Hz and were connected through an asynchronous pipeline to reduce blocking, resulting in an end-to-end teleoperation latency below $0.1\,\mathrm{s}$ (mean). This low-latency feedback was critical for contact-sensitive detachment, improving operator stability and the overall quality of the collected demonstrations.

\subsection{Data Analysis}
The following dataset statistics were reported: 3.71~hours total duration, 227 episodes, 491 effective picking attempts (including both successes and failures), 40~s average duration per pick, and 1.5 average retries per pick. These statistics characterized the long-horizon nature of the dataset.

To quantify diversity, episodes were grouped by greenhouse region and collection batch, and coverage across illumination conditions (e.g., backlit/side-lit/low-light), occlusion levels (leaf occlusion, branch interference, strawberries overlap), and target maturity was summarized. Such coverage analysis helped identify the main sources of generalization error and domain shift.

Table~\ref{tab:dataset_coverage} summarizes the coverage ratio (by episodes) of several key factors in our dataset. Category definitions followed consistent criteria: illumination conditions were determined by scene brightness and shadow ratio; occlusion levels were determined by the visible-area ratio of the target fruit and the degree of leaf occlusion; and scene complexity was approximated by the number of visible fruits.

\begin{table}[t]
\centering
\caption{Dataset coverage ratios by key factors (episode-level).}
\label{tab:dataset_coverage}
\begin{tabularx}{\columnwidth}{c c >{\centering\arraybackslash}X}
\toprule
Factor & Category & Ratio (\% episodes) \\
\hline
\multirow{3}{*}{Illumination} & Low light & 18 \\
 & Normal & 62 \\
 & Strong specular & 20 \\\hline
\multirow{3}{*}{Occlusion} & Mild & 60 \\
 & Moderate & 25 \\
 & Heavy & 15 \\\hline
\multirow{3}{*}{Visible targets} & 1 & 62 \\
 & 2 & 30 \\
 & $\geq$3 & 8 \\
\hline
\multirow{2}{*}{Target maturity} & Red only & 71 \\
 & Mixed & 29 \\
\bottomrule
\end{tabularx}
\end{table}

Unlike highly constrained ``clean'' manipulation datasets, teleoperation trajectories were intentionally not over-restricted. Within safety constraints, operators were allowed to choose search order, approach routes, end-effector orientations, and detachment motions (e.g., gently pushing leaves, approaching around obstacles, and varying bending tempo). Importantly, natural failure-and-recovery segments were retained (e.g., re-localization after failed approach, second attempts after slip, and correction after failed placement) rather than discarded, which better matched the distribution encountered in closed-loop deployment.

\subsection{Policy Adaptation}
\label{sec:policy_adaptation}
We chose three representative open-source VLA models as baselines, covering different pretraining scales and action representations: $\pi_{0}$ \cite{pi0_2024}, $\pi_{0.5}$ \cite{pi05_2025}, and Wall-oss \cite{walloss_2025}. Unless otherwise specified, all models were adapted under the same data split and training budget.

At each control step $t$, the policy outputted a continuous arm command $a_t^{\mathrm{arm}}\in\mathbb{R}^7$, which was executed by the low-level controller in velocity mode. In addition, it outputted a discrete pump command for the compliant gripper, $a_t^{\mathrm{pump}}\in\{\mathrm{in},\mathrm{out},\mathrm{idle}\}$, corresponding to suction, inflation, and idle. The final action vector was $a_t=[a_t^{\mathrm{arm}},a_t^{\mathrm{pump}}]$.

Let the observation at time $t$ be $o_t=(\{I_t^{(v)}\}_{v=1}^{3}, s_t)$, where $\{I_t^{(v)}\}_{v=1}^{3}$ are RGB images from the three cameras and $s_t\in\mathbb{R}^8$ is the robot state. Given $o_t$ and the language goal $l$, the policy predicted $a_t$. We used the demonstrated action $\hat{a}_t$ as supervision and optimize: 
\begin{equation}
\mathcal{L}=\lambda_{\mathrm{arm}}\cdot \|a_t^{\mathrm{arm}}-\hat{a}_t^{\mathrm{arm}}\|_2^2+\lambda_{\mathrm{pump}}\cdot \mathrm{CE}(a_t^{\mathrm{pump}},\hat{a}_t^{\mathrm{pump}}),
\label{eq:policy_loss}
\end{equation}
where, $\mathrm{CE}(\cdot)$ denotes cross-entropy loss, and $\lambda_{\mathrm{arm}}$ and $\lambda_{\mathrm{pump}}$ balance the gradient scales of continuous arm control and discrete actuator commands.

Two fine-tuning strategies were considered:
(1) Full fine-tuning. All model parameters were updated to maximize adaptation to the new environment and embodiment. The best performance under this setting was reported and used as an upper bound for parameter-efficient methods.
(2) Parameter-efficient fine-tuning via LoRA \cite{hu2021loralowrankadaptationlarge}. To reduce compute and mitigate overfitting, low-rank adapters were injected into selected linear projections of the attention and MLP blocks, with only the LoRA \cite{hu2021loralowrankadaptationlarge} parameters trained while keeping the backbone weights frozen. We use the default configuration provided by openpi (commit: 981483dca0fd9acba698fea00aa6e52d56a66c58) and wall-x (commit: 97406f2ab5de414c79b091873f946c112d105c72). In Sec.~\textup{III}, LoRA \cite{hu2021loralowrankadaptationlarge} and full fine-tuning were compared in terms of success rate, damage rate, and efficiency.

All models were trained using 2$\times$A800 (80~GB) GPUs with batch size 16 for 6 epochs. Training time and loss were reported in Table~\ref{tab:lora_full_compare}.

\begin{table}[!t]
  \centering
  \caption{Loss and training time of LoRA vs.\ full fine-tuning on different models.}
  \label{tab:lora_full_compare}
  \resizebox{\columnwidth}{!}{%
  \begin{tabular}{lcccccc}
    \toprule
    \multirow{2}{*}{Method} &
    \multicolumn{2}{c}{$\pi_0$} &
    \multicolumn{2}{c}{$\pi_{0.5}$} &
    \multicolumn{2}{c}{wall-oss} \\
    \cmidrule(lr){2-3}\cmidrule(lr){4-5}\cmidrule(lr){6-7}
    & loss & time & loss & time & loss & time \\
    \midrule
    lora & 0.00241 & 39h52m12s & 0.00329 & 43h18m04s & 0.00211 & 42h46m29s \\
    full & 0.02702 & 42h47m53s & 0.02981 & 45h16m38s & 0.01758 & 47h32m51s \\
    \bottomrule
  \end{tabular}%
  }
\end{table}

\subsection{Deployment}
All models were deployed locally on an NVIDIA RTX A5000 GPU (16~GB). Two inference and control deployment modes were considered:

(1) Synchronous inference. The system ran in a serial loop of ``image acquisition $\rightarrow$ model inference $\rightarrow$ action execution''. The control loop waited for inference to finish before entering the next cycle. When the inference latency $t_{\mathrm{inf}}$ approached or exceeded the control period $1/f_c$, the effective action rate dropped and control jitter could occur, potentially causing overshoot, slip, or missed contact windows during contact-sensitive detachment.

(2) Asynchronous inference. To reduce control jitter, an asynchronous two-thread pipeline was adopted:
(i) Inference thread: when the action queue fell below a threshold, the latest observation was read and inference was triggered. The policy output an action chunk, which was converted into timestamped $\{\mathrm{TimedAction}\}$ entries and appended to a shared action queue.
(ii) Real-time control (RTC) thread: following the LeRobot RTC \cite{cadene2024lerobot}, the control thread ran at a fixed frequency $f_c$ (30~Hz in this work). At each cycle, it popped the next action for the current timestep and sent it to the robot, ensuring that the control-time reference was governed by the control thread (instead of the inference thread).

\begin{figure}[t]
  \centering
  \includegraphics[width=\linewidth]{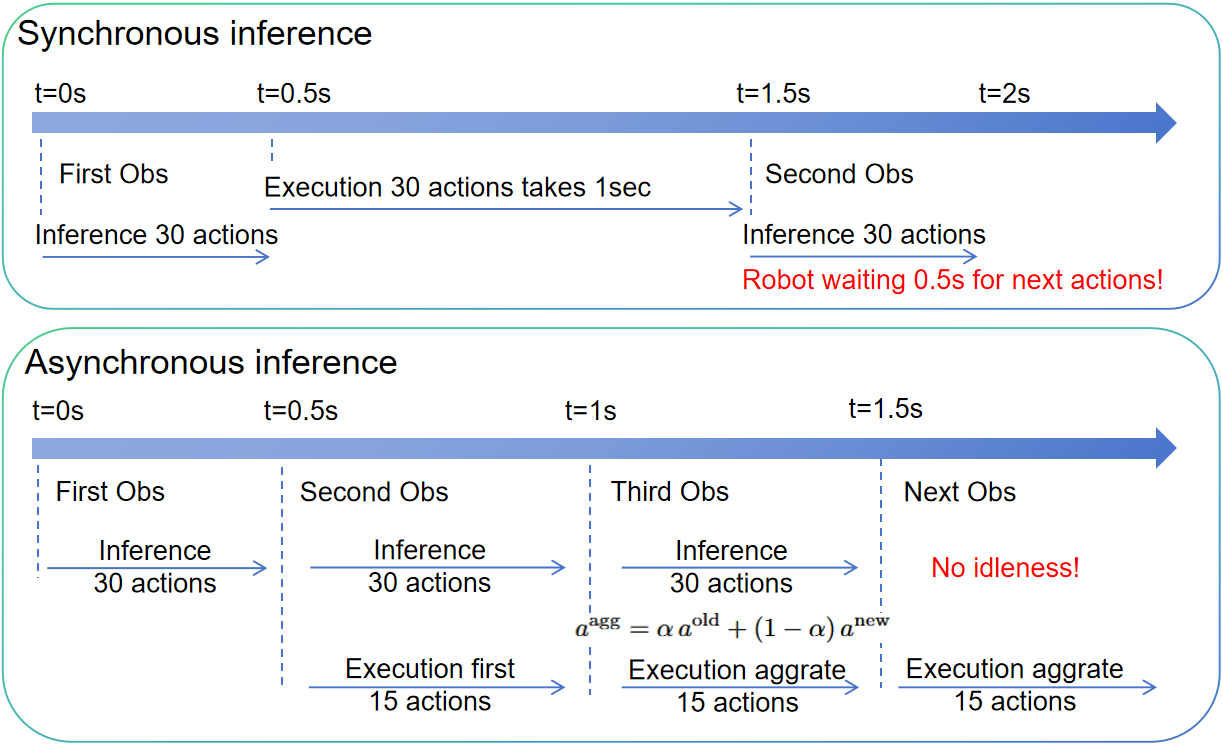}  
  \caption{Flow diagram showing the processing pipelines for synchronous (blocking, sequential) and asynchronous (non-blocking, concurrent) inference modes.}
  \label{fig:inference_pipeline}
\end{figure}

The processing pipelines for synchronous and asynchronous inference modes were illustrated in Fig.~\ref{fig:inference_pipeline}. Since inference latency could reach hundreds of milliseconds, chunked action prediction was used to amortize inference cost, together with queue-threshold triggering: a new inference was requested only when the remaining actions in the queue fell below $\tau$ (50\% of one chunk), balancing responsiveness and stability.

When a new action chunk arrived, its steps could temporally overlap with the tail of the existing queue. To avoid abrupt changes and ensure smooth transitions, the overlapping segment was combined via weighted averaging:
\begin{equation}
a^{\mathrm{agg}}=\alpha\, a^{\mathrm{old}}+(1-\alpha)\, a^{\mathrm{new}},
\label{eq:action_aggregation}
\end{equation}
where, $a^{\mathrm{old}}$ denotes the not-yet-executed actions in the queue, $a^{\mathrm{new}}$ denotes the newly inferred actions, and $\alpha\in[0,1]$ controlled the smoothness. The aggregation was applied to the continuous arm command, while the discrete pump command followed the newest action.

To prevent the robot from stalling when the queue was unexpectedly depleted, a \emph{must-go} event was introduced: once the control thread detected an empty action queue, it immediately forced the inference thread to run and replenish the queue, ensuring continuous closed-loop execution. In Sec.~\textup{III}, synchronous and asynchronous deployment modes were further compared in terms of evaluation metrics.

\section{EXPERIMENTS}
\subsection{Experimental Setup}
All experiments were conducted in a commercial tabletop strawberry farm (Cuihu Agriculture Technology Co., Ltd., Beijing, China), with an equal number of tests conducted in sunny and overcast conditions and at different times of day. For reproducibility, the control frequency was fixed to 30~Hz and all camera images were resized to $640\times 480$ RGB inputs. For each model or method, 50 trials were conducted under the same protocol. For all VLA methods, a unified language prompt template was used: ``Pick all ripe strawberries and place them into the tray.''

Evaluations were performed across fruits with random maturity and challenging conditions (e.g., leaf occlusions, branch interference, and fruit overlap). Different methods were compared under the same dataset and protocol. Unless otherwise specified, asynchronous inference was used.

\begin{table*}[!h]
\centering
\scriptsize
\caption{Metrics for different models under different fine-tuning strategies and training levels; 2, 4, and 6 denote the number of training epochs.}
\setlength{\tabcolsep}{4pt}
\begin{tabularx}{\textwidth}{l l *{9}{>{\centering\arraybackslash}X} >{\centering\arraybackslash}X}
\toprule
\multirow{2}{*}{Fine-tuning} & \multirow{2}{*}{Metrics}
  & \multicolumn{3}{c}{PI0} & \multicolumn{3}{c}{wall-oss} & \multicolumn{3}{c}{PI05}& \multirow{2}{*}{Best} \\
\cmidrule(lr){3-5}\cmidrule(lr){6-8}\cmidrule(lr){9-11}
 & & 2~(ep) & 4~(ep) & 6~(ep) & 2~(ep) & 4~(ep) & 6~(ep) & 2~(ep) & 4~(ep) & 6~(ep) & \\
\midrule
\multirow{5}{*}{Full}
 & SS (score) & 20.4 & 46.6 & 72.6 & 26.4 & 52.4 & 78.8 & 30.8 & 58.4 & 82.6 & 82.6 \\
 & SR (\%) & 14.0 & 38.0 & 60.0 & 24.0 & 44.0 & 68.0 & 30.0 & 50.0 & 74.0 & 74.0 \\
 & Cycle time (s) & 53.6 & 42.1 & 38.4 & 53.3 & 52.8 & 46.3 & 44.2 & 40.7 & 32.6 & 32.6 \\
 & DR (\%) & 5.2 & 4.1 & 4.2 & 3.5 & 4.0 & 3.9 & 4.4 & 3.8 & 4.1 & 3.5 \\
\midrule
\multirow{5}{*}{LoRA}
 & SS (score) & 18.6 & 42.4 & 68.2 & 22.8 & 48.2 & 72.8 & 26.8 & 50.4 & 73.6 & 73.6 \\
 & SR (\%) & 12.0 & 34.0 & 54.0 & 20.0 & 38.0 & 60.0 & 25.0 & 42.0 & 64.0 & 64.0 \\
 & Cycle time (s) & 55.6 & 40.7 & 38.2 & 55.2 & 50.8 & 45.2 & 46.2 & 39.6 & 38.3 & 38.2 \\
 & DR (\%) & 5.0 & 4.8 & 4.9 & 4.9 & 5.2 & 4.4 & 4.7 & 3.6 & 3.8 & 3.6 \\
\bottomrule
\end{tabularx}
\label{tab:results}
\end{table*}

\subsection{Metrics}
Strawberry-harvesting performance was evaluated from three perspectives: task completion, operational efficiency, and fruit quality.

Success Score (SS). To quantify partial progress in long-horizon episodes, a stage-based success score was calculated following the task decomposition in Sec.~\textup{II}.A. Each episode was divided into $K$ stages ($K=5$ in this work). For stage $k$, a binary completion indicator $c_k\in\{0,1\}$ was assigned. The per-episode success score was defined as:
\begin{equation}
\mathrm{SS}_{i}=\sum_{k=1}^{K} w_k\, c_k,
\label{eq:success_score}\end{equation}
where, $w_k\ge 0$ were stage weights that satisfied $\sum_{k=1}^{K} w_k=1$. Uniform weights were used by default, i.e., $w_k=1/K$. Over $N$ evaluation trials ($N=50$ in this work), the final score was reported on a scale of 0--100 by averaging scores per-episode and multiplying by 100 (note that $\mathrm{SS}\in[0,1]$ by construction):

\begin{equation}
\mathrm{SS}=100\cdot \frac{1}{N}\sum_{i=1}^{N} \mathrm{SS}_i.
\label{eq:success_score_100}
\end{equation}

For alignment with prior work, the binary Success Rate (SR) was also reported:
\begin{equation}
\mathrm{SR}=\frac{N_{\mathrm{succ}}}{N_{\mathrm{total}}},
\label{eq:success_rate}
\end{equation}
where $N_{\mathrm{succ}}$ and $N_{\mathrm{total}}$ denote the number of successful and total trials.

Efficiency. We report efficiency using first-attempt cycle time, defined as the average duration from episode start to a successful pick and placement conditioned on success on the first attempt (s/pick), reflecting the nominal speed of a single closed-loop execution without retries.

Damage Rate (DR). To capture both the existence and severity of damage, each successfully placed strawberry $i$ was annotated with a damage severity score $s_i\in\{0,1,2,3,4,5\}$, where, $s_i=0$ indicates no observable damage and higher values corresponded to more severe defects under predefined criteria (e.g., bruising, skin breakage, juice leakage, and obvious deformation), capped at 5. The severity was normalized as $\hat{s}_i=s_i/5$, and the following metric was defined:
\begin{equation}
\mathrm{DR}=\frac{1}{N_{\mathrm{succ}}}\sum_{i=1}^{N_{\mathrm{succ}}}\hat{s}_i.
\label{eq:damage_rate}
\end{equation}

\subsection{Main Results}
The main results were summarized in Table~\ref{tab:results}. Success rate (SR), cycle time, and damage rate (DR) were reported. Across models, both full fine-tuning and LoRA \cite{hu2021loralowrankadaptationlarge} improved SR as the number of training epochs increased (2 to 6), while cycle time generally decreased, indicating more efficient execution with additional training. Full fine-tuning consistently achieved higher SR and SS than LoRA \cite{hu2021loralowrankadaptationlarge} at the same epoch count, with comparable damage rates, suggesting that performance gains primarily stemmed from improved task completion rather than increased fruit damage. 

Among the evaluated models, $\pi_{0.5}$ \cite{pi05_2025} achieved the highest SR (74\%) while maintaining a low damage rate (4.1\%), suggesting more reliable long-horizon closed-loop execution.

\begin{figure*}[t]
  \centering
  \subfloat[success]{\includegraphics[width=0.85\textwidth]{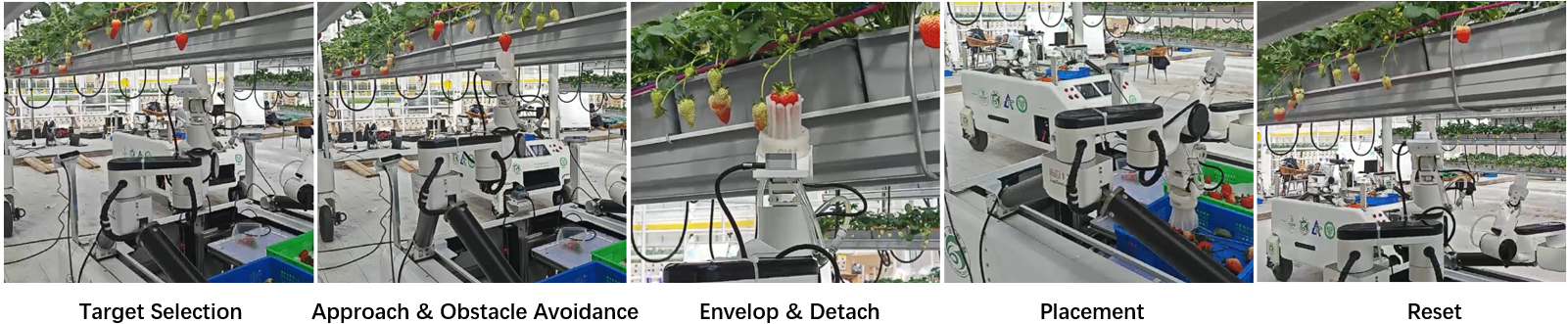}}\par\medskip
  \subfloat[fail\_no\_inhalation]{\includegraphics[width=0.85\textwidth]{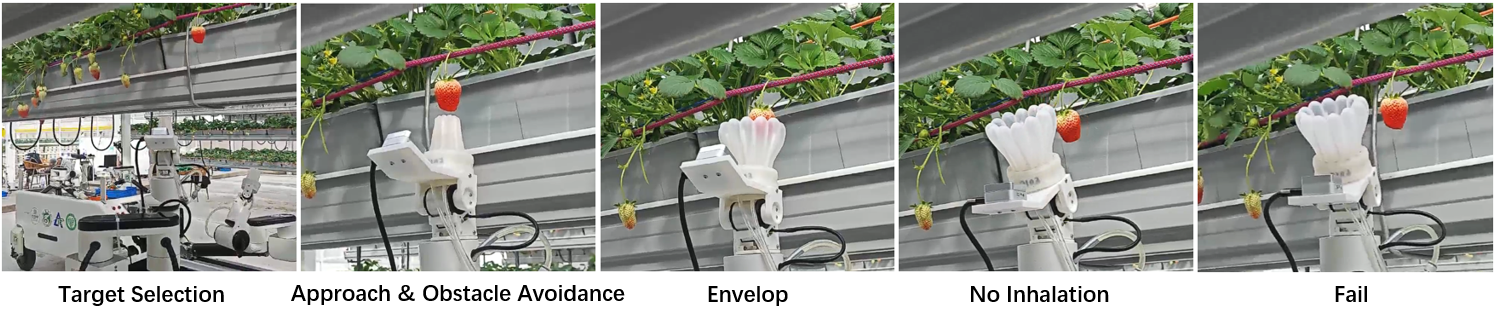}}\par\medskip
  \subfloat[fail\_wrist\_block]{\includegraphics[width=0.85\textwidth]{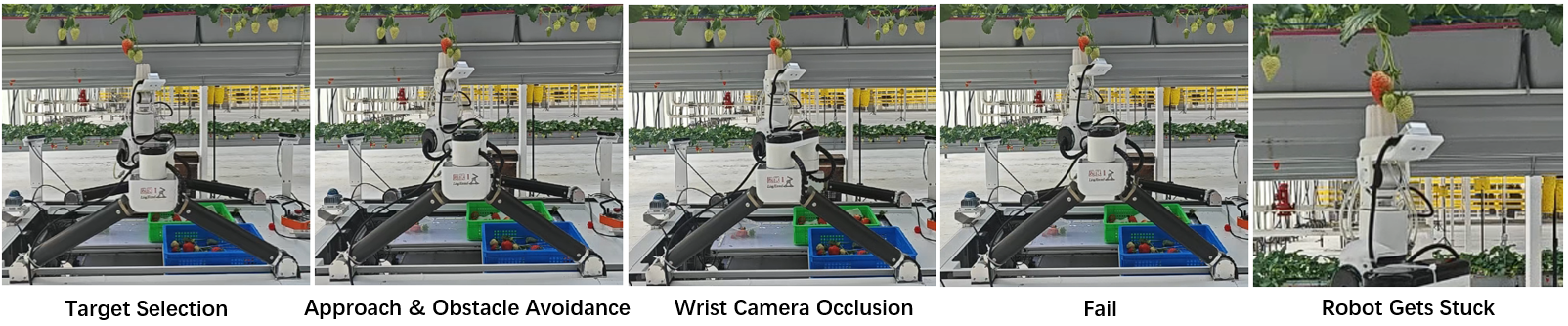}}\par\medskip
  \subfloat[retry\_for\_failures]{\includegraphics[width=0.85\textwidth]{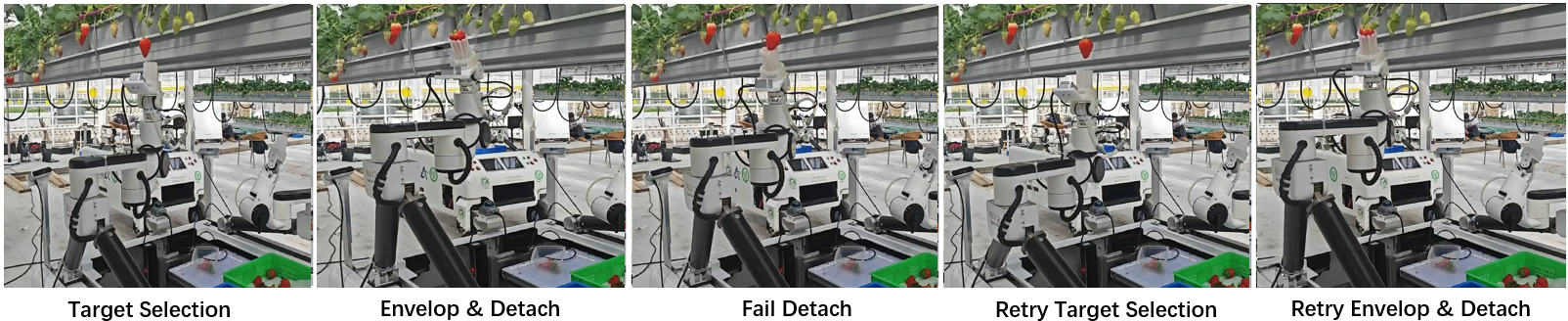}}
  \caption{Key frames showing representative successful,  failed and retry cases.}
  \label{fig:failures}
\end{figure*}

Fig.~\ref{fig:failures} (a) shows a representative successful trial, while Fig.~\ref{fig:failures} (b) and (c) show representative failure cases. VLA policies were able to execute the full closed-loop pipeline with reasonable robustness; however, the failures shown in the figure included: (1) the gripper inflated and enveloped the strawberry, but the fruit rotated without suction, so detachment failed and the episode later became stuck; and (2) severe occlusions from the wrist view during approach prevented the policy from executing the subsequent action sequence. Additional failure cases included inability to localize ripe strawberries and align beneath the fruit, as well as repeated slippage that led to a human-labeled failure; these cases were rare and thus not shown.

Inference deployment was further investigated using $\pi_{0.5}$ \cite{pi05_2025}, the best-performing baseline, with results summarized in Table~\ref{tab:sync_async}. Although synchronous and asynchronous deployments achieved comparable coarse target localization in most trials, synchronous inference remained tightly coupled to inference latency and the control period: a reduced action update rate or output jitter could induce overshoot, slip, or missed contact windows during detachment, lowering overall detachment success. In contrast, asynchronous inference decoupled perception--inference from low-level control, allowing the controller to execute the latest available commands at a stable frequency and reducing frame-blocking-induced jitter, thereby improving stability in the contact-sensitive envelop-and-detach stage.

\begin{table}[t]
\centering
\caption{Comparison of synchronous and asynchronous inference ($\pi_{0.5}$).}
\label{tab:sync_async}
\begin{tabularx}{\columnwidth}{p{0.25\columnwidth} >{\centering\arraybackslash}X >{\centering\arraybackslash}X}
\toprule
Metric & Sync & Async \\
\midrule
SS (score)            & 76.2 & 82.6 \\
SR (\%)               & 70.0   & 74.0   \\
Cycle time (s)        & 45.7 & 32.6 \\
DR (\%)               & 3.8  & 4.1  \\
\bottomrule
\end{tabularx}
\end{table}

\subsection{Ablations}
Controlled ablation studies were conducted to quantify the impact of camera-view configurations on closed-loop performance. Three settings were compared: (1) a single scene camera, (2) two scene cameras, and (3) two scene cameras combined with a wrist-mounted camera (the default multi-view configuration). The results were reported in Table~\ref{tab:camera_views}.

\begin{table}[t]
\centering
\caption{Impact of camera-view configurations on performance.}
\label{tab:camera_views}
\begin{tabularx}{\columnwidth}{l *{3}{>{\centering\arraybackslash}X}}
\toprule
Metric & Left/Right & Left+Right & All \\
\hline
SS (score) & 16.4 & 42.4 & 82.6 \\
SR (\%) & 10.0 & 42.0 & 74.0 \\
Cycle time (s) & 75.3 & 61.3 & 32.6 \\
DR (\%) & 5.5 & 4.3 & 4.1 \\
\bottomrule
\end{tabularx}
\end{table}

The empirical findings were consistent with the intended functional roles of the sensors. Moving from Left/Right to Left+Right yielded clear gains in SS and SR (16.4$\rightarrow$42.4, 10\%\,$\rightarrow$\,42\%), with shorter cycle time (0.2$\rightarrow$0.6 fruit, 75.3$\rightarrow$61.3\,s), indicating that wider scene context improved target search and coarse localization. Adding the wrist camera produced the largest improvement across metrics: SS reached 82.6 and SR 74\%, with cycle time dropping to 32.6\,s. The DR changed only slightly (5.5\%$\rightarrow$4.1\%); the small increase observed in some trials was mainly attributable to repeated grasp attempts, which introduced additional contact events. These results suggested that close-range, end-effector-aligned observations were critical for contact-sensitive stages such as enveloping and detachment, where local geometry cues and fine relative pose determined slippage and successful separation.

\subsection{Comparisons}
To more comprehensively characterize the advantages and limitations of VLA policies for real-world agricultural picking, performance was compared against a conventional modular baseline and analyzed from three perspectives: robustness to occlusion and reflections, damage control, and picking efficiency. The quantitative comparison was summarized in Table~\ref{tab:conv_vs_vla}.

\begin{table}[t]
\centering
\caption{Comparison between conventional (Conv) and VLA pipelines.}
\label{tab:conv_vs_vla}
\begin{tabularx}{\columnwidth}{l *{2}{>{\centering\arraybackslash}X}}
\toprule 
Metrics & Conv & VLA \\
\hline
SS (score) & 91.8 & 82.6 \\
SR (\%) & 89.0 & 74.0 \\
Cycle time (s) & 8.3 & 32.6 \\
DR (\%) & 3.8 & 4.1 \\
\bottomrule
\end{tabularx}
\end{table}

Robustness to occlusion and reflections. The modular pipeline was often bottlenecked by perception failures under severe occlusions and specular reflections, which subsequently caused planning and servoing failures and could require substantial engineering effort and model retraining. In contrast, VLA policies directly consumed RGB observations and could exploit multi-view context to localize strawberries more accurately, especially during search and coarse approach. The lower SR in Table~\ref{tab:conv_vs_vla} (82.6/74\% vs.\ 91.8/89\%) was mainly driven by failures in subsequent contact-rich stages (enveloping and detachment), where close-range observability loss, compliance mismatch, and action timing became dominant error sources rather than initial target localization.

Damage rate. In our setup, the compliant end-effector mitigates most contact-induced damage for both methods. Table~\ref{tab:conv_vs_vla} shows a small gap (3.8\% vs. 4.1\%), consistent with occasional repeated attempts and longer interaction time for VLA.

Picking efficiency. The conventional pipeline typically achieved higher throughput and shorter cycle time (5.5 fruit, 8.3\,s) than VLA (1.5 fruit, 32.6\,s) in Table~\ref{tab:conv_vs_vla}. Its control loop was lightweight and could run faster with a stable control frequency. End-to-end VLA deployment required real-time perception and inference, inference latency reduced the effective control rate and introduced control intermittency, thereby increasing the average cycle time. 

Failure handling. When detachment failed, the conventional system often continued executing the remaining routine in a fixed order---attempting placement, resetting, and then re-localizing for a new try---even though the strawberry had not been harvested. In contrast, VLA policies typically persisted on the current subtask and continued attempting grasping and detachment instead of skipping ahead, as shown in Fig.~\ref{fig:failures} (d), which better matched the closed-loop objective.

\subsection{Discussion}
Despite lower metric values, VLA offered a clear advantage in development cost and time: conventional pipelines typically demanded a multidisciplinary team to design, tune, and integrate perception, planning, and control over long development cycles, whereas VLA policies could often be adapted by a single developer in a much shorter time. In this work, fewer than four hours of real-world demonstrations already yielded non-trivial success, underscoring the potential of VLA for rapid deployment in agricultural harvesting.


\section{CONCLUSIONS}
This paper investigated transferring VLA policies to real-world tabletop strawberry harvesting, a long-horizon and contact-sensitive task in cluttered greenhouse environments. An end-to-end closed-loop system was presented that combined multi-view RGB sensing, VR teleoperation for long-horizon demonstrations, and a unified training--deployment pipeline to execute the full picking cycle. To the best of our knowledge, this study provided the first systematic real-robot validation of VLA policies for strawberry harvesting in greenhouse tabletop settings. Experiments under a unified real-greenhouse protocol showed that adaptation on real demonstrations was essential, and that deployment-level inference--control decoupling improved stability for contact-rich interaction: asynchronous inference yielded higher completion and efficiency, particularly during the detachment stage. Among the evaluated baselines, fully fine-tuned $\pi_{0.5}$ achieved the best overall performance, while sensor ablations highlighted the importance of wrist-centric close-range observations for reliable stage transitions under occlusion and specular reflections. The remaining limitations stemmed from degraded observability under severe occlusions/reflections, contact-dynamics mismatch, and insufficient hard-case data coverage. Future work will expand diverse real-world data, enhance end-effector-centric sensing, and optimize low-latency deployment and action interfaces to reduce cycle time while maintaining fruit quality and safety. 

 




\bibliographystyle{IEEEtran}
\nocite{10.1016/j.compag.2024.109468}
\nocite{Xiong2020}
\nocite{rajendran2023towards}
\nocite{ahn2022saycan}
\nocite{sapkota2025vision}
\nocite{chen2024design}
\nocite{11231358}
\nocite{cadene2024lerobot}
\nocite{xiao2024review}
\nocite{zhu2025_yolov11_skp}
\nocite{sobol2024_headgrabber}
\nocite{ji2022_contact_impedance}
\nocite{sun2026srrnet}
\nocite{brohan2023_rt2}
\nocite{yang2025_agriGPTvl}
\nocite{zhao2024_vlmpc}
\nocite{ge2025_occlusion_aware_sequence}
\nocite{kim2024_openvla}
\nocite{sapkota2026visionlanguageactionvlamodelsconcepts}
\nocite{pi0_2024}
\nocite{pi05_2025}
\nocite{walloss_2025}
\nocite{hu2021loralowrankadaptationlarge}
\bibliography{references}

@article{10.1016/j.compag.2024.109468,
author = {Guo, Jie and Yang, Zhou and Karkee, Manoj and Jiang, Qianjing and Feng, Xuping and He, Yong},
title = {Technology progress in mechanical harvest of fresh market strawberries},
year = {2024},
issue_date = {Nov 2024},
publisher = {Elsevier Science Publishers B. V.},
address = {NLD},
volume = {226},
number = {C},
issn = {0168-1699},
url = {https://doi.org/10.1016/j.compag.2024.109468},
doi = {10.1016/j.compag.2024.109468},
journal = {Comput. Electron. Agric.},
month = nov,
numpages = {47}
}

@article{Xiong2020,
  author = {Xiong, Ya and Ge, Yuanyue and Grimstad, Lars and From, P{\aa}l J.},
  title = {An autonomous strawberry-harvesting robot: Design, development, integration, and field evaluation},
  journal = {Journal of Field Robotics},
  year = {2020},
  volume = {37},
  number = {2},
  pages = {202--224},
  doi = {10.1002/rob.21889}
}

@article{rajendran2023towards,
  title = {Towards Autonomous Selective Harvesting: A Review of Robot Perception, Robot Design, Motion Planning and Control},
  author = {Rajendran, Vishnu and Debnath, Bappaditya and Mghames, Sariah and Mandil, Willow and Parsa, Soran and Parsons, Simon and Ghalamzan-E, Amir},
  journal = {arXiv preprint arXiv:2304.09617},
  year = {2023},
  url = {https://arxiv.org/abs/2304.09617}
}

@article{ahn2022saycan,
  title = {Do As I Can, Not As I Say: Grounding Language in Robotic Affordances},
  author = {Ahn, Michael and Brohan, Anthony and Brown, Noah and Chebotar, Yevgen and Cortes, Omar and ... and Levine, Sergey}, 
  journal = {arXiv preprint arXiv:2204.01691},
  year = {2022},
  url = {https://arxiv.org/abs/2204.01691}
}

@article{sapkota2025vision,
  title={Vision-Language-Action (VLA) Models: Concepts, Progress, Applications and Challenges},
  author={Sapkota, Ranjan and Cao, Yang and Roumeliotis, Konstantinos I and Karkee, Manoj},
  journal={arXiv preprint arXiv:2505.04769},
  year={2025}
}

@inproceedings{chen2024design,
  title={Design and control of a novel six-degree-of-freedom hybrid robotic arm},
  author={Chen, Yang and Miao, Zhonghua and Ge, Yuanyue and Lin, Sen and Chen, Liping and Xiong, Ya},
  booktitle={2024 IEEE/RSJ International Conference on Intelligent Robots and Systems (IROS)},
  pages={3597--3604},
  year={2024},
  organization={IEEE}
}

@ARTICLE{11231358,
  author={Yang, Lichao and Li, Haitao and Lin, Sen and Xiong, Ya},
  journal={IEEE Robotics and Automation Letters}, 
  title={High-Tolerance Soft-Rigid Gripper for Low-Damage Robotic Strawberry Harvesting}, 
  year={2026},
  volume={11},
  number={1},
  pages={121-128},
  doi={10.1109/LRA.2025.3630528}
}

@misc{cadene2024lerobot,
    author = {Cadene, Remi and Alibert, Simon and Soare, Alexander and Gallouedec, Quentin and Zouitine, Adil and Palma, Steven and Kooijmans, Pepijn and Aractingi, Michel and Shukor, Mustafa and Aubakirova, Dana and Russi, Martino and Capuano, Francesco and Pascal, Caroline and Choghari, Jade and Moss, Jess and Wolf, Thomas},
    title = {LeRobot: State-of-the-art Machine Learning for Real-World Robotics in Pytorch},
    howpublished = "\url{https://github.com/huggingface/lerobot}",
    year = {2024}
}

@article{xiao2024review,
  title = {Review of Research Advances in Fruit and Vegetable Harvesting Robots},
  author = {Xiao, Xu and Wang, Yaonan and Jiang, Yiming},
  journal = {Journal of Electrical Engineering \& Technology},
  volume = {19},
  number = {1},
  pages = {773--789},
  year = {2024},
  doi = {10.1007/s42835-023-01596-8},
  url = {https://doi.org/10.1007/s42835-023-01596-8}
}

@article{zhu2025_yolov11_skp,
  author = {Zhu, Tianxiao and Zhang, Wei and Miao, Zhonghua and Zhao, Chunjiang and Xiong, Ya},
  title = {YOLOv11-SKP: an enhanced model for strawberry bounding box and key point detection in harvesting scenarios},
  journal = {Engineering Research Express},
  year = {2025},
  volume = {7},
  number = {4},
  pages = {045217},
  doi = {10.1088/2631-8695/ae0de3},
  url = {https://doi.org/10.1088/2631-8695/ae0de3},
  note = {published 14 October 2025; access: 2026-02-21}
}

@article{sobol2024_headgrabber,
  author = {Sobol, Zygmunt and Kurpaska, S\l{}awomir and Nawara, Piotr and others},
  title = {Prototype of a New Head Grabber for Robotic Strawberry Harvesting with a Vision System},
  journal = {Sensors (Basel)},
  year = {2024},
  volume = {24},
  number = {20},
  pages = {6628},
  doi = {10.3390/s24206628},
  url = {https://doi.org/10.3390/s24206628}
}

@article{ji2022_contact_impedance,
  author = {Ji, Wei and Tang, Chencheng and Xu, Bo and He, Guozhi},
  title = {Contact force modeling and variable damping impedance control of apple harvesting robot},
  journal = {Computers and Electronics in Agriculture},
  year = {2022},
  volume = {198},
  pages = {107026},
  doi = {10.1016/j.compag.2022.107026},
  url = {https://doi.org/10.1016/j.compag.2022.107026}
}

@article{sun2026srrnet,
  title        = {Vision-Based Early Fault Diagnosis and Self-Recovery for Strawberry Harvesting Robots},
  author       = {Sun, Meili and Zhao, Chunjiang and Yang, Lichao and Liu, Hao and Hu, Shimin and Xiong, Ya},
  journal      = {arXiv preprint arXiv:2601.02085},
  year         = {2026},
  url          = {https://arxiv.org/abs/2601.02085}
}

@article{brohan2023_rt2,
  author = {Brohan, Anthony and Brown, Noah and Carbajal, Justice and Chebotar, Yevgen and Chen, Xi and Choromanski, Krzysztof and ...},
  title = {RT-2: Vision-Language-Action Models Transfer Web Knowledge to Robotic Control},
  journal = {arXiv preprint arXiv:2307.15818},
  year = {2023},
  url = {https://arxiv.org/abs/2307.15818}
}

@article{yang2025_agriGPTvl,
  author = {Yang, Bowen and Chen, ...},
  title = {AgriGPT-VL: Agricultural Vision–Language Understanding Suite},
  journal = {arXiv preprint arXiv:2510.04002},
  year = {2025},
  url = {https://arxiv.org/abs/2510.04002}
}

@inproceedings{zhao2024_vlmpc,
  author = {Zhao, Wentao and Chen, Jiaming and Meng, Ziyu and Mao, Donghui and Song, Ran and Zhang, Wei},
  title = {VLMPC: Vision-Language Model Predictive Control for Robotic Manipulation},
  booktitle = {Robotics: Science and Systems (RSS) / proceedings},
  year = {2024},
  url = {https://ppjmchen.github.io/vlmpc/} 
}

@misc{ge2025_occlusion_aware_sequence,
  author = {Ge, Yuanyue and Zhao, Quan and Sun, Meili},
  title = {Occlusion-Aware Harvesting Sequence Planning for Strawberries in Clusters Using Large Language and Vision-Language Models},
  howpublished = {SSRN preprint},
  year = {2025},
  doi = {10.2139/ssrn.5553835},
  url = {https://doi.org/10.2139/ssrn.5553835},
  note = {preprint; accessed: 2026-02-21}
}

@article{kim2024_openvla,
  author = {Kim, Moo Jin and Pertsch, Karl and ... and others},
  title = {OpenVLA: An Open-Source Vision-Language-Action Model},
  journal = {arXiv preprint arXiv:2406.09246},
  year = {2024},
  url = {https://arxiv.org/abs/2406.09246}
}

@misc{sapkota2026visionlanguageactionvlamodelsconcepts,
      title={Vision-Language-Action (VLA) Models: Concepts, Progress, Applications and Challenges}, 
      author={Ranjan Sapkota and Yang Cao and Konstantinos I. Roumeliotis and Manoj Karkee},
      year={2026},
      eprint={2505.04769},
      archivePrefix={arXiv},
      primaryClass={cs.CV},
      url={https://arxiv.org/abs/2505.04769}, 
}

@misc{pi0_2024,
  title = {$\pi_0$: A Vision-Language-Action Flow Model for General Robot Control},
  author = {{Physical Intelligence (Pi) Team}},  year = {2024},
  howpublished = {arXiv preprint arXiv:2410.24164},
  url = {https://arxiv.org/abs/2410.24164}
}

@misc{pi05_2025,
  title = {$\pi_{0.5}$: a Vision-Language-Action Model with Co-Training for Open-World Generalization},
  author = {{Physical Intelligence (Pi) Team}},
  year = {2025},
  howpublished = {arXiv preprint arXiv:2504.16054},
  url = {https://arxiv.org/abs/2504.16054}
}

@misc{walloss_2025,
  title = {WALL-OSS: Igniting VLMs toward the Embodied Space},
  author = {X-Square Robot / WALL Team},
  year = {2025},
  howpublished = {arXiv preprint arXiv:2509.11766},
  url = {https://arxiv.org/abs/2509.11766}
}

@misc{hu2021loralowrankadaptationlarge,
      title={LoRA: Low-Rank Adaptation of Large Language Models}, 
      author={Edward J. Hu and Yelong Shen and Phillip Wallis and Zeyuan Allen-Zhu and Yuanzhi Li and Shean Wang and Lu Wang and Weizhu Chen},
      year={2021},
      eprint={2106.09685},
      archivePrefix={arXiv},
      primaryClass={cs.CL},
      url={https://arxiv.org/abs/2106.09685}, 
}

\end{document}